\documentclass[11pt]{article}

\usepackage[preprint]{acl}

\usepackage{times}
\usepackage{latexsym}

\usepackage[T1]{fontenc}

\usepackage[utf8]{inputenc}

\usepackage{microtype}

\usepackage{inconsolata}

\usepackage{graphicx}
\usepackage{booktabs}
\usepackage{wrapfig}

\usepackage[algo2e,ruled]{algorithm2e}
\newif\ifdrafting
\draftingtrue 
\ifdrafting

\else

\fi

\definecolor{guppiegreen}{rgb}{0.0, 0.7, 0.7}
\definecolor{alizarin}{rgb}{0.8, 0.1, 0.2}
\definecolor{brandeisblue}{rgb}{0.0, 0.44, 1.0}

\usepackage{amsmath}
\usepackage{multirow}

\usepackage{tabularx}
\usepackage{array}

\newcolumntype{Y}{>{\raggedright\arraybackslash}X}
\usepackage{amssymb}

\usepackage[table]{xcolor}

\definecolor{methodbg}{HTML}{F7EEE6}

\usepackage{booktabs}
\usepackage{multirow}
\usepackage{amssymb}
\usepackage[table]{xcolor}

\definecolor{cachebg}{RGB}{242,246,250}
\newcommand{\groupsep}{\arrayrulecolor{black!25}\midrule[0.3pt]\arrayrulecolor{black}}

\newcommand{\method}{TeCoP}
\title{TeCoP: Accelerating Diffusion Large Language Model Decoding via Tentative Context Pipelining}

\author{%
  \textbf{Jui-Chien Tsou}$^{*}$ \quad
  \textbf{Hsuan-Chi Liu}$^{*}$ \quad
  \textbf{Wei-Chin Wang}$^{*}$\\
  \textbf{Yu-Syuan Xu} \quad
  \textbf{Chun-Yi Lee}\\
  \\
  National Taiwan University%
}

\begin{document}
\maketitle

\begingroup
\renewcommand{\thefootnote}{*}
\footnotetext{Equal contribution.}
\endgroup

\begin{abstract}
Diffusion large language models (dLLMs) predict multiple token positions per forward pass, yet block-wise decoders serialize generation and repeatedly recompute stable predictions. We attribute this inefficiency to coupling a prediction's readiness for irreversible commitment with its utility as context. We introduce Tentative Context Pipelining (\method{}), a training-free decoder that decouples these roles through staggered block-parallel pipelining and draft-and-verify. \method{} reuses stable but uncommitted predictions as provisional cross-block context, with commitment governed by confidence and stability checks. Across four LLaDA and Dream dLLMs on GSM8K, Minerva MATH, and MBPP, \method{} reduces number of forward evaluations by up to 87\% and achieves up to 7.50$\times$ throughput without KV caching and 11.77$\times$ with caching over the uncached original decoder. It outperforms Fast-dLLM and DSB in throughput across all 12 model--benchmark settings, with and without caching, while keeping average cached accuracy within one percentage point of the original decoder.
\end{abstract}
\section{Introduction}
\label{sec:intro}

The decoding efficiency of diffusion large language models (dLLMs) hinges on how aggressively their parallel prediction capacity is converted into committed tokens. Unlike autoregressive models that generate strictly left to right, dLLMs iteratively refine masked positions through denoising~\citep{he2023diffusionbert,sahoo2024simple} and can predict many positions in a single forward pass. Recent large-scale instances such as LLaDA~\citep{nie2026large} and Dream~\citep{ye2025dream} reach competitive accuracy on general language understanding, instruction following, and reasoning benchmarks. This parallel potential, however, remains largely unrealized in practice, and the bottleneck lies in how generation is currently scheduled. Existing decoders~\citep{arriola2025block,wu2025fastdllm} pair block-wise semi-autoregressive schedules with conservative confidence-based commit rules: a later block waits until its predecessors are resolved, and within the active block only the positions whose confidence clears a high threshold are committed, even though every masked position is predicted in the same forward pass. The rest are masked again and recomputed on the next pass, so each pass advances only a handful of tokens and decoding degenerates toward the token-by-token generation that diffusion models were designed to escape. As a result, dLLMs spend much of their compute re-deciding already stable predictions, leaving their parallel prediction capacity underused.

\begin{figure}[t]
    \centering
    \includegraphics[width=0.95\linewidth]{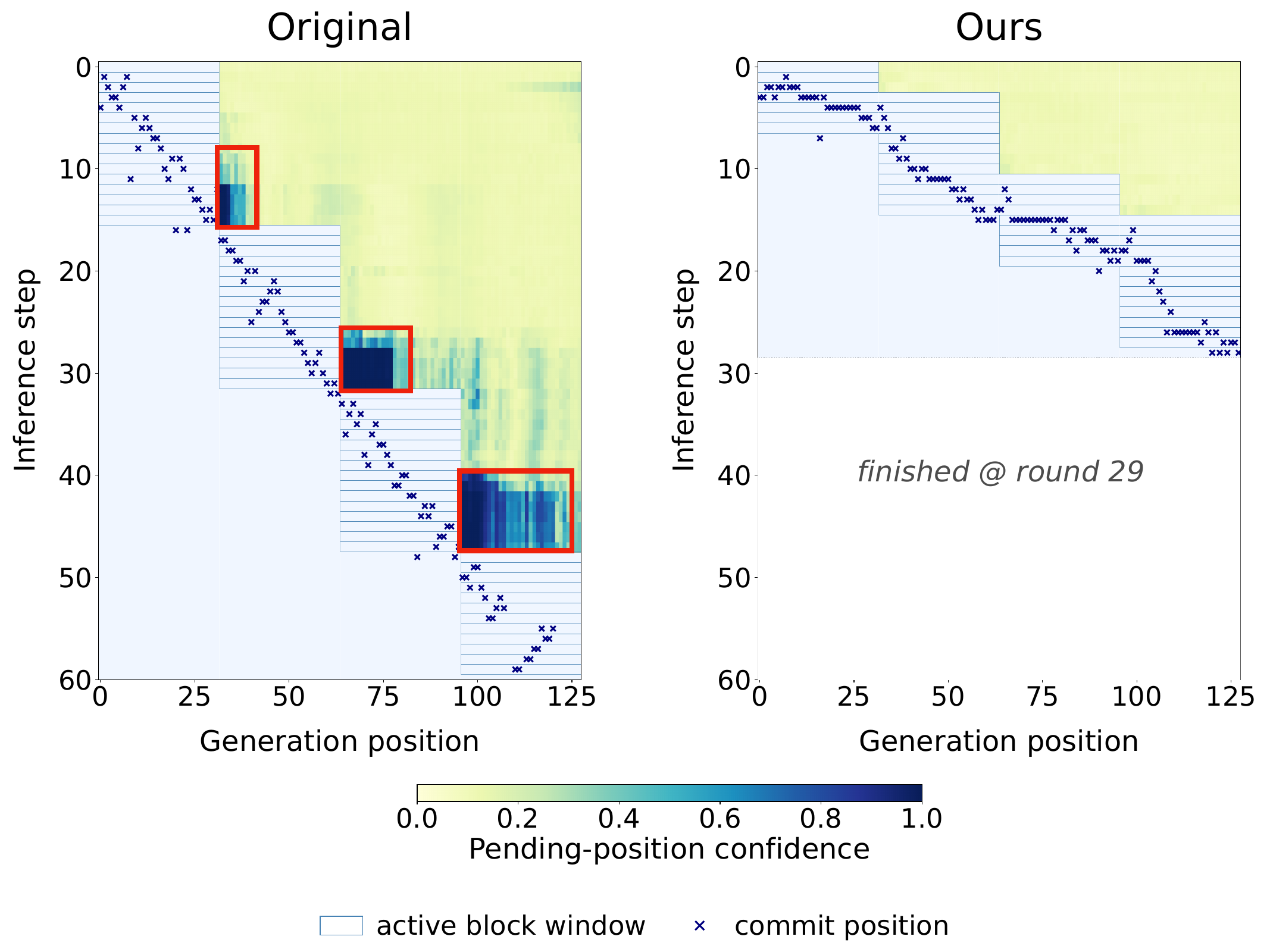}
    \caption{\textbf{Visualization of pending-position confidence during masked diffusion decoding.} \textbf{Left:} In the baseline, many positions outside the active block window achieve high confidence but cannot be committed. 
    \textbf{Right:} Our pipeline method progressively expands the active block region, enabling these predictions to be committed earlier and reducing wasted forward passes. 
    }
    \label{fig:teaser_pipeline}
    \vspace{-10pt}
\end{figure}

Recent work has sought faster dLLM decoding through approximate caching, confidence-thresholded parallel commitment, and adaptive block schedules~\citep{wu2025fastdllm,israel2025apd,mohamed2025sched,luo2026dsb}. Much of it rests on a shared observation that a prediction often stabilizes several steps before it crosses the commit threshold, and confidence-only commitment is therefore conservative~\citep{wang2025creditdecoding,li2025prophet,zhang2026leap,luo2026dawn}. These methods act on that signal either by committing such tokens earlier or by deferring uncertain ones, and in both cases the predictions stay outside the decoding context until finalized. The caution is deliberate, as an unreliable prediction admitted into the context could propagate its error to every position that conditions on it. The cross-block utility of uncommitted predictions nevertheless remains unexploited, since a future-block token can become predictable before its preceding block resolves, while conventional block-wise decoding masks these positions and defers later blocks. The one pipeline that crosses block boundaries, D2F~\citep{wang2025d2f}, obtains this ability by distilling the model to condition on partially denoised prefixes, and no training-free counterpart exists for frozen models. Beyond this cross-block loss, the same masking discards stable predictions within a block that have yet to clear the commit threshold, and forces the redundant recomputation noted above. Across these methods, an uncommitted prediction is rarely treated as more than a candidate to commit or discard, and two gaps open between what a model has already produced and what it is permitted to exploit as context, one across blocks and one within them.

Both gaps identified above trace to a single root cause: existing decoders treat two distinct properties of a prediction as one. A prediction has a readiness to be committed, which requires it to be reliable enough to finalize irreversibly, and a utility as context, which requires only that it be accurate enough to guide other positions. Since a prediction enters the working context only once it clears the commit threshold, the decoder structurally binds the second property to the first. Fig.~\ref{fig:teaser_pipeline} illustrates the resulting waste: under a conventional schedule, many positions outside the active block reach high confidence well before they become eligible to commit, yet remain masked and repeatedly recomputed. Separating these properties allows stable but uncommitted predictions to serve as provisional context for other blocks. Realizing this requires two components: a schedule that keeps several blocks open and a verification mechanism that checks tentative predictions before commitment.


\begin{figure*}[t]
    \centering
    \includegraphics[width=\linewidth]{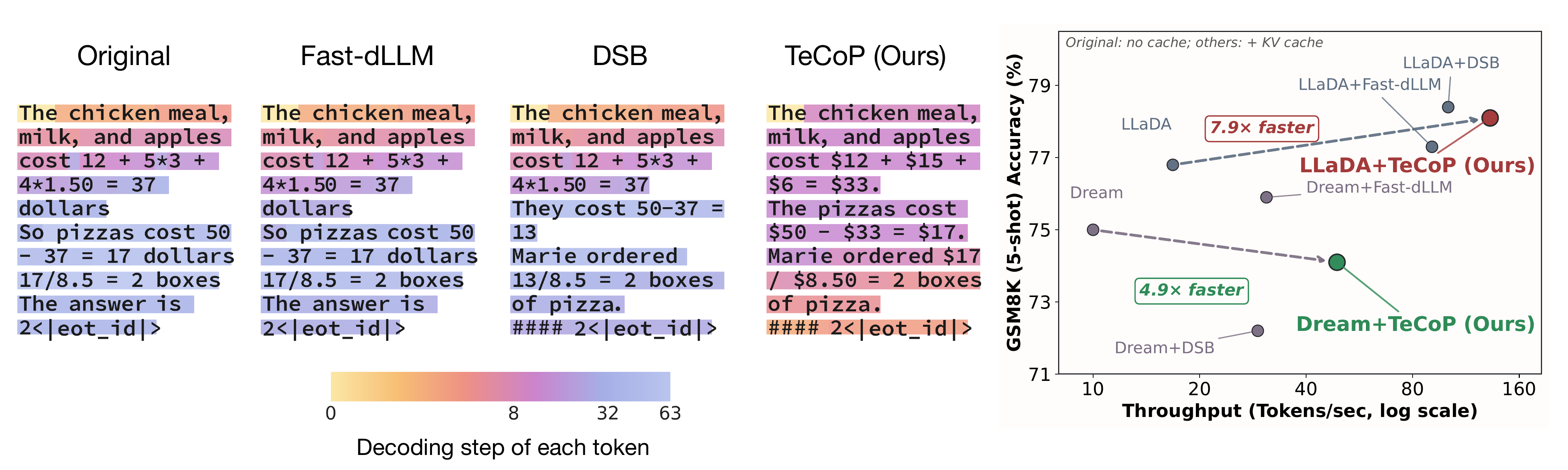}
    \caption{\textbf{Overview of the decoding efficiency of Tentative Context Pipelining (\method{}).}
    \textbf{Left:} Decoding trajectories of the original block-wise decoder, Fast-dLLM~\citep{wu2025fastdllm}, DSB~\citep{luo2026dsb}, and \method{}, colored by the step at which each token is committed. \method{} activates multiple blocks in parallel and finishes in fewer number of 
    forward evaluations.
    \textbf{Right:} Throughput--accuracy trade-off on GSM8K with KV caching; speedups are relative to the uncached original decoder.}
    \label{fig:teaser}
    \vspace{-10pt}
\end{figure*}

To this end, we propose Tentative Context Pipelining (\method{}), a training-free decoder that requires no change to the model. It realizes the schedule as a block-parallel pipeline that keeps multiple blocks active within a single forward pass, and the verification discipline as a draft-and-verify procedure that reuses uncommitted-but-stable predictions as provisional cross-block context, while commitment remains governed by confidence and cross-round stability. Fig.~\ref{fig:teaser} illustrates the resulting decoding trajectories and throughput--accuracy trade-off. Our contributions are as follows:




\begin{itemize}
  \item We identify the coupling between commitment readiness and contextual utility as a key bottleneck in dLLM decoding, which can be relaxed at inference time without retraining.

  \item We propose \method{}, a training-free decoder that combines cross-block pipelining with the reuse of stable but uncommitted predictions as provisional context, committing them only after verification.

  \item Across four LLaDA and Dream dLLMs on GSM8K, Minerva MATH, and MBPP, \method{} outperforms Fast-dLLM and DSB in throughput across all settings, achieving 1.03--1.31$\times$ Fast-dLLM's throughput without KV caching and 1.26--1.58$\times$ with caching. Relative to standard block-wise decoding, it reaches up to 7.50$\times$ and 11.77$\times$ speedups without and with caching, respectively, while keeping average accuracy within one percentage point.
\end{itemize}



\section{Related Work}

\paragraph{Diffusion Language Models.}
Diffusion language models cast text generation as iterative denoising over discrete tokens, and masked diffusion has become the dominant instantiation of this paradigm~\citep{he2023diffusionbert,sahoo2024simple}. The paradigm has since expanded into a broad family of systems: open-weight models such as LLaDA and Dream address general instruction following and mathematical reasoning~\citep{nie2026large,ye2025dream}, while Gemini Diffusion, Mercury, and Seed Diffusion demonstrate the approach in industrial settings~\citep{google2025geminidiffusion,khanna2025mercury,song2025seed}. Dream-Coder and DiffuCoder further specialize this line for code generation~\citep{xie2025dreamcoder,gong2025diffucoder}. Within this family, our work builds on masked dLLMs: the model architecture, tokenizer, and training objective remain unchanged, and the improvement lies entirely in inference-time decoding.

\paragraph{Block-wise Decoding for Diffusion Language Models.}
Block-wise semi-autoregressive decoding is the prevailing inference strategy for dLLMs: the output is split into contiguous blocks, and each is decoded from the committed tokens of earlier blocks~\citep{arriola2025block}. This coarse left-to-right structure improves reliability and forms the basis of most efficient dLLM decoders. Existing refinements target different parts of this pipeline, through confidence-based and adaptive commitment~\citep{wu2025fastdllm,israel2025apd,mohamed2025sched,wang2025creditdecoding,chen2025ccd,li2025prophet,zhang2026leap}, dependency-aware token selection~\citep{luo2026dawn}, cache reuse across denoising steps~\citep{wu2025fastdllmv2,ma2025dkvcache,liu2025dllmcache,jiang2025d2cache}, and dynamic block schedules~\citep{luo2026dsb,mohamed2025sched}. These refinements largely treat decoding as a selection problem over positions, thresholds, or schedules. Since their blocks still run in sequence, the decoder advances only within a narrow active region and leaves the parallel predictions already computed for later regions underused.

\paragraph{Accelerating Sequential Block-wise Decoding.}
A growing body of work relaxes the strict sequencing of block-wise decoding. Among these, the method most relevant to ours is Discrete Diffusion Forcing~\citep{wang2025d2f}, which pipelines blocks under staggered activation thresholds and conditions later blocks on partially denoised earlier ones. It reaches this inter-block parallelism through asymmetric distillation that retrains the model to denoise from partial prefixes, whereas \method{} adds no training and runs on frozen dLLMs. The same relaxation appears in other cross-block and streaming schemes that decode later regions before earlier blocks are fully resolved~\citep{liu2025wedlm,wu2025fastdllmv2}, and in learned parallel decoders that distill faster certainty dynamics into the model~\citep{chen2025dparallel,bao2025learn2pd}. Beyond these diffusion-specific schedules, sequential dependence has also been reduced at the algorithm level: Jacobi and lookahead decoding solve for several tokens at once~\citep{santilli2023parallel,fu2024lookahead}, and speculative decoding proposes draft tokens and verifies them against the target model~\citep{leviathan2023speculative,chen2023speculative}, an idea recently carried into diffusion generation~\citep{christopher2025speculative,li2025diffuspec,chen2026dflash,wu2025freedave,gao2025ssd,cui2026simsd}. Nearest in spirit to \method{} is a line that treats non-final predictions as usable signal, though each method exploits it differently: Soft-Mask represents uncertain predictions as soft tokens~\citep{hersche2025softmask}, WINO verifies or remasks unreliable ones~\citep{hong2026wide}, and deferred-commitment schedules wait on uncertain positions for more context~\citep{shu2026dcd,luo2026dawn}. \method{} reuses this signal in a different way: selected uncommitted predictions are retained as discrete tentative drafts inside the same masked-dLLM state, supply provisional context across active blocks, and are committed only through a confidence or cross-round stability check. Unlike the trained pipeline of~\citet{wang2025d2f}, \method{} obtains this cross-block context at inference time on a frozen model. Unlike methods that defer or discard uncertain tokens, it reuses stable predictions as context before they commit.

\section{Preliminaries}
\subsection{Masked Diffusion Language Models}

\method{} builds on the decoding process of masked diffusion language models, and this section sets up the decoding loop and notation that the method later modifies. In these models, generation reverses a forward process that masks tokens in a sequence. Specifically, at timestep $s$ each position of a clean sequence $x_0 = (x_{0,1}, \ldots, x_{0,L})$ is either preserved or replaced by the $[\mathrm{MASK}]$ symbol, as follows:


\begin{equation}
x_{s,i}
=
\begin{cases}
x_{0,i}, & \text{with probability } \rho_s, \\
[\mathrm{MASK}], & \text{with probability } 1-\rho_s .
\end{cases}
\end{equation}
where $\rho_s$ is the probability that a token is preserved. The model $p_\theta$ is trained to recover the original token at each masked position, formulated as:
\begin{equation}
p_\theta(x_{0,i} \mid x_s), \quad i \in \mathcal{M}_s,
\end{equation}
with $\mathcal{M}_s = \{ i \mid x_{s,i} = [\mathrm{MASK}] \}$ the set of masked positions. At inference, decoding no longer tracks the diffusion timestep and instead proceeds in rounds indexed by $r$. Each round performs one forward pass and then commits a subset of positions, and the working sequence $x^{(r)}$ holds committed tokens in place while the rest stay masked. Standard block-wise decoding feeds this sequence back unchanged, with $z^{(r)} = x^{(r)}$ as the model input. From $z^{(r)}$, every masked position is scored in parallel over the vocabulary $\mathcal{V}$, and each yields a most likely token and a confidence, computed as:

\begin{equation}
\begin{aligned}
\hat{x}^{(r)}_i
&= \arg\max_{v \in \mathcal{V}}
p_\theta(x_i = v \mid z^{(r)}), \\
c^{(r)}_i
&= \max_{v \in \mathcal{V}}
p_\theta(x_i = v \mid z^{(r)}).
\end{aligned}
\end{equation}

A single forward pass therefore scores every masked position at once, and how much of this parallel capacity a schedule turns into committed tokens is what separates one decoder from another.

\subsection{Block-wise Diffusion Decoding}

This section formalizes the schedule that turns the model's per-round predictions into a finished sequence: the block-wise, left-to-right loop that almost every dLLM decoder adopts, and the one \method{} later reorganizes. Such a decoder splits the response into $m = \lceil L/b \rceil$ contiguous blocks of length $b$, defined as follows:
\begin{equation}
B_j = \{(j-1)b+1, \ldots, \min(jb, L)\}.
\end{equation}
These blocks are decoded from left to right, and a block is completed before the next opens. At each round, completed blocks act as fixed context, the active block is decoded in parallel, and later blocks remain masked. Within the active block $B_j$, every uncommitted position is scored in one forward pass, and a confidence gate commits the positions whose confidence reaches a threshold $\tau$, given by:
\begin{equation}
\mathcal{C}^{(r)}
=
\left\{
i \in B_j
\;\middle|\;
x^{(r)}_i = [\mathrm{MASK}],\;
c^{(r)}_i \geq \tau
\right\}.
\end{equation}
Committed tokens are fixed and never revised, and the rest stay masked for a later round. This schedule trades parallelism for stability by deferring uncertain predictions, and its cost is that a later block cannot contribute until the blocks before it are resolved. Relaxing this constraint is what \method{} sets out to do.

\begin{figure*}[!t]
    \centering
    \includegraphics[width=\linewidth]{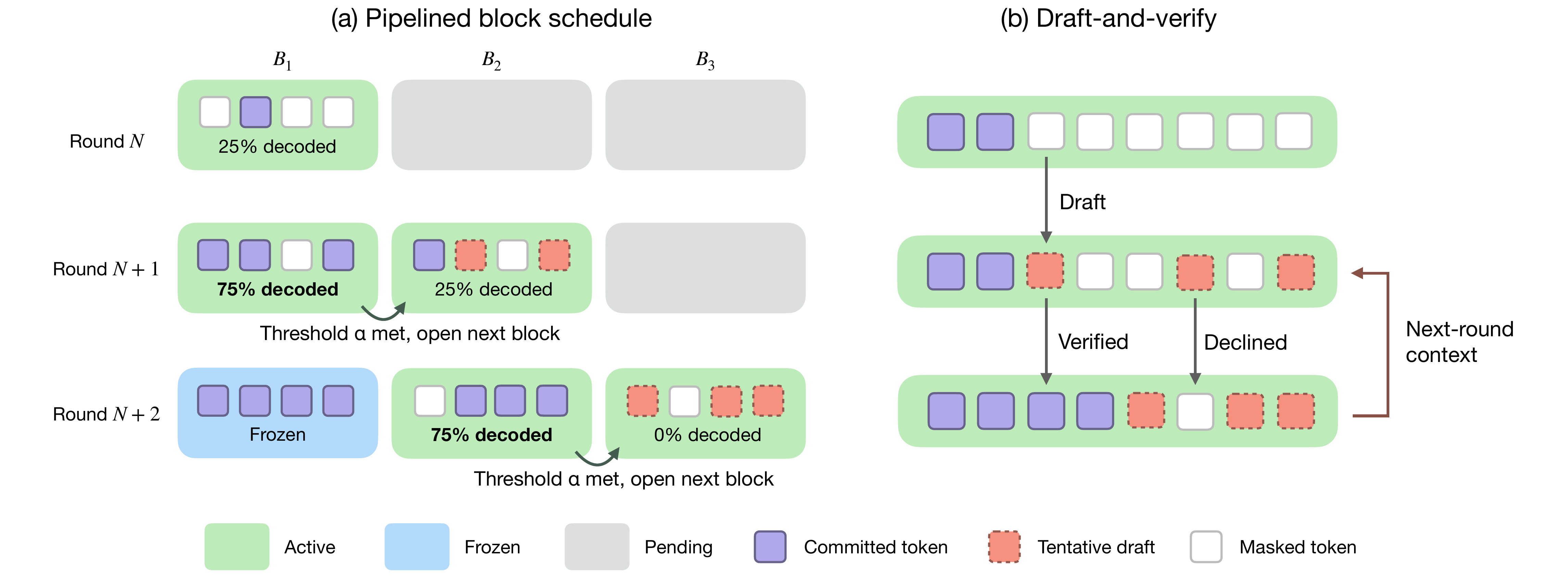}
    \caption{\textbf{Illustration of Tentative Context Pipelining (\method{}) mechanism.} The pipelined block schedule keeps multiple blocks simultaneously active and progressively opens new blocks according to the activation threshold $\alpha$ (set to 75\% here for illustration). The draft-and-verify procedure recycles uncommitted predictions as tentative drafts, uses them as provisional context in the next round, and commits them through a confidence or stability check.}
    \label{fig:method_pipeline}
\end{figure*}

\section{Methodology}
\label{sec:method}

This section presents the architecture of \method{}, which realizes the decoupling identified in Section~\ref{sec:intro}: a prediction may guide other positions as context long before it is reliable enough to commit. Fig.~\ref{fig:method_pipeline} illustrates the full design, built from two mechanisms that share one pool of active blocks. The \textbf{block-parallel pipeline schedule} advances several blocks at once and allows a later block to begin before its predecessor finishes. The \textbf{draft-and-verify procedure} reuses uncommitted-but-stable predictions as provisional context to help other unresolved positions decode, while using stability as an additional signal for commitment.  Algorithm~\ref{alg:tecop} states the complete loop.

\begin{algorithm2e}[!t]
\footnotesize
\DontPrintSemicolon
\caption{\method{} Decoding}
\label{alg:tecop}

\KwIn{prompt, length $L$, block size $b$, depth $K$,
thresholds $\tau,\tau_d,\tau_v,\alpha$}
\KwOut{committed sequence $x$}

$x \leftarrow [\mathrm{MASK}] \times L$;\quad
$\tilde{x} \leftarrow \varnothing \times L$
\tcp*{tentative drafts}

$\{B_1,\dots,B_m\} \leftarrow \mathrm{split}(L,b)$;\quad
$\mathrm{activate}(B_1)$\;

\While{not $\mathrm{all\_frozen}(B_1,\dots,B_m)$}{

  $\mathcal{A} \leftarrow$ positions of active blocks\;

  \tcp{build input with tentative context}
  $z \leftarrow x$\;
  \For{$i\in\mathcal{A}$ with $x_i=[\mathrm{MASK}]$}{
    \lIf{$\tilde{x}_i \ne \varnothing$}{
      $z_i \leftarrow \tilde{x}_i$
    }
  }

  \tcp{one forward pass; scores every masked $i\in\mathcal{A}$}
  $\hat{x}_i \leftarrow \arg\max_{v\in\mathcal{V}} p_\theta(x_i{=}v \mid z)$;\quad
  $c_i \leftarrow \max_{v\in\mathcal{V}} p_\theta(x_i{=}v \mid z)$\;

  $n \leftarrow \mathrm{num\_committed}(x)$\;

  \tcp{safe-commit or verify-commit}
  \For{$i\in\mathcal{A}$ with $x_i=[\mathrm{MASK}]$}{
    \uIf{$c_i \ge \tau$}{
      $x_i \leftarrow \hat{x}_i$
      \tcp*{safe}
    }
    \ElseIf{
      $\tilde{x}_i \ne \varnothing$
      \textbf{and} $\hat{x}_i=\tilde{x}_i$
      \textbf{and} $c_i\ge\tau_v$
    }{
      $x_i \leftarrow \hat{x}_i$
      \tcp*{verified}
    }
  }

  \tcp{ensure progress if no token commits}
  \If{$\mathrm{num\_committed}(x)=n$}{
      $i^\star \leftarrow
      \arg\max_{i\in\mathcal{A}:\,x_i=[\mathrm{MASK}]} c_i$\;
      $x_{i^\star} \leftarrow \hat{x}_{i^\star}$\;
  }

  \tcp{refresh tentative drafts}
  \For{$i\in\mathcal{A}$ with $x_i=[\mathrm{MASK}]$}{
    \lIf{$\tilde{x}_i\ne\varnothing$ \textbf{or} $c_i\ge\tau_d$}{
      $\tilde{x}_i \leftarrow \hat{x}_i$
    }
  }

  \tcp{advance pipeline}
  \For{$B_j$ in active blocks}{
    \lIf{$\mathrm{fully\_committed}(B_j)$}{
      $\mathrm{freeze}(B_j)$
    }
  }
  \If{$\mathrm{committed\_ratio}(\mathrm{frontier})\ge\alpha$
      \textbf{and} $\mathrm{num\_active}()<K$}{
    $\mathrm{activate}(\mathrm{next\_pending}())$\;
  }
}

\Return $x$\;

\end{algorithm2e}

\subsection{Pipelined Block Schedule}
\label{sec:pipeline}

\method{} achieves its concurrency through the pipeline schedule, which maintains a sliding window of active blocks in place of the single block of standard decoding. The window is bounded on two axes. The pipeline depth $K$ caps its width, the largest number of blocks active together. The activation threshold $\alpha$ controls its advance: the next block opens once the current frontier block has committed an $\alpha$-fraction of its tokens. As this window slides across the sequence, each block passes through three states: \textit{pending} while it waits fully masked, \textit{active} once it enters the window and joins each forward pass, and \textit{frozen} after all its tokens commit and it becomes fixed context. The activation threshold $\alpha$ protects quality as the window advances. Although the parallel forward pass could open all $K$ blocks at once, premature activation carries a cost: a block opened before its predecessor commits anything must predict from an almost fully masked left context, and its unreliable predictions propagate through the rounds that follow. The threshold rules this out by opening a block only after the current frontier block has committed an $\alpha$-fraction of its tokens, which anchors every new block in reliable context. Under this schedule, later blocks advance while their predecessors are still resolving, and this recovers the parallelism that strict left-to-right decoding forgoes.

\subsection{Drafting with Provisional Context}

\begin{figure}[t]
\centering


\begingroup
\footnotesize
\setlength{\tabcolsep}{2.5pt}
\renewcommand{\arraystretch}{1.15}

\captionof{table}{States in an active block.}
\label{tab:posstates}

\vspace{2pt}

\begin{tabular}{
    @{}
    p{0.25\columnwidth}
    c
    p{0.50\columnwidth}
    @{}
}
\toprule
State & Symbol & Meaning \\
\midrule
Committed
& $x_i$
& finalized; value locked \\

Tentative
& $\tilde{x}_i$
& draft from a prior round \\

Masked
& \texttt{[M]}
& no draft yet \\
\bottomrule
\end{tabular}

\endgroup

\vspace{8pt}


\begin{minipage}[t]{0.485\columnwidth}
    \centering
    \includegraphics[
        width=\linewidth
    ]{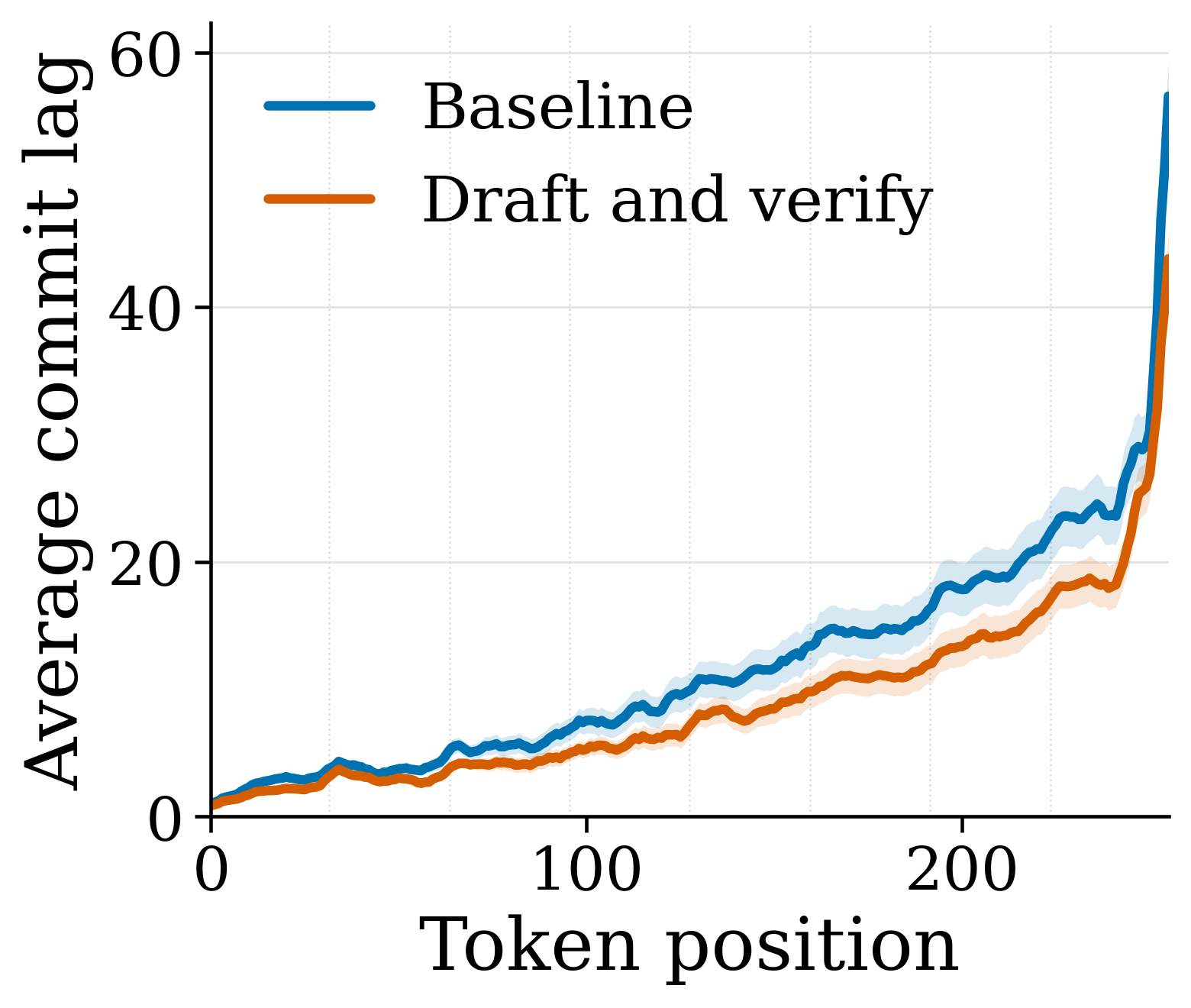}
\end{minipage}
\hfill
\begin{minipage}[t]{0.485\columnwidth}
    \centering
    \includegraphics[
        width=\linewidth
    ]{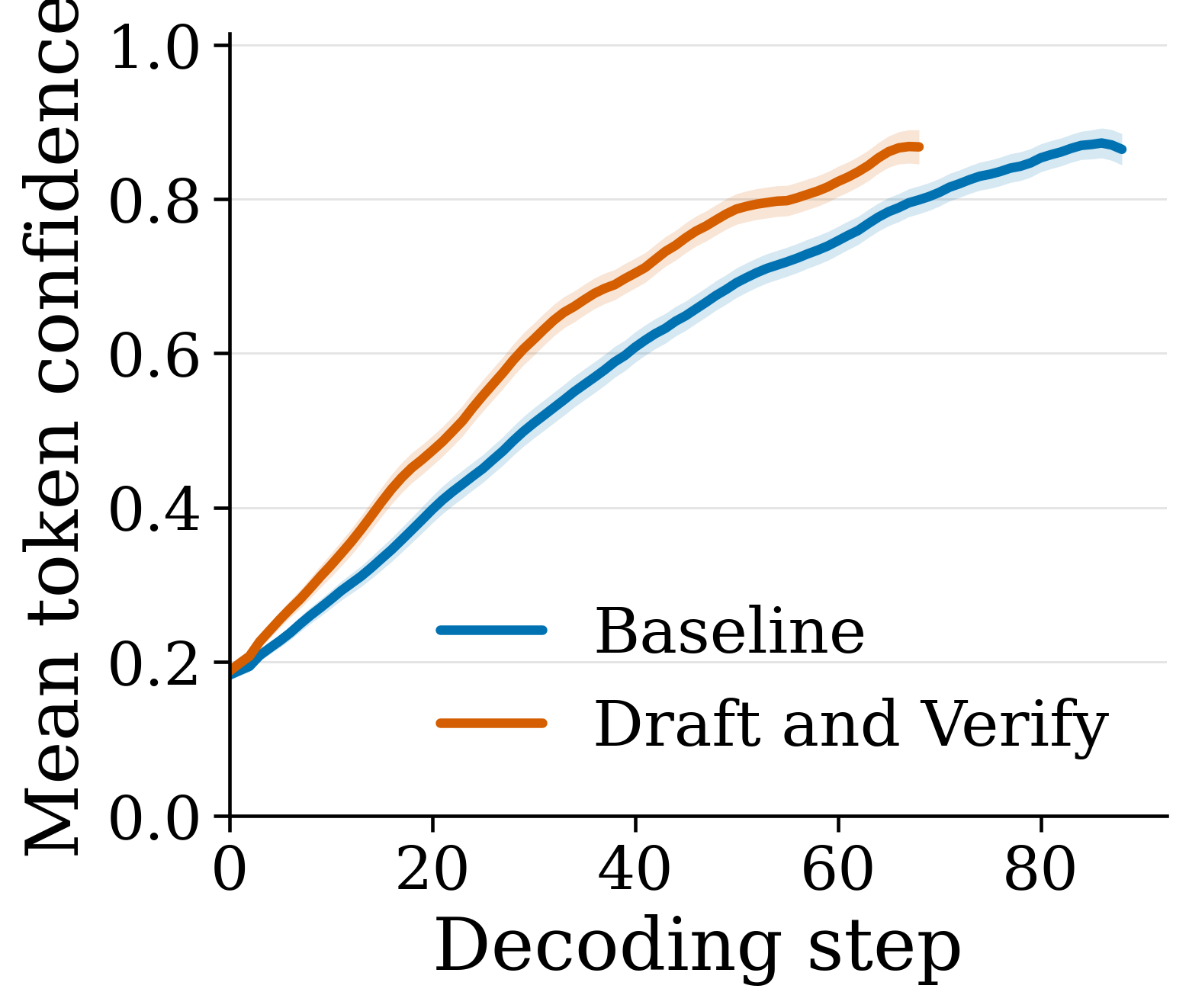}
\end{minipage}

\vspace{-3pt}

\caption{
Draft-and-verify analysis on 100 GSM8K samples.
Left: commit lag by token position.
Right: token confidence over decoding steps.
}
\label{fig:draft_analysis}

\end{figure}

With several blocks overlapping, the draft-and-verify procedure makes use of the predictions they produce before those predictions are committed. A position whose confidence clears a draft threshold $\tau_d$ becomes a \textit{tentative} draft: it serves as provisional context in the next round and may later be committed if it remains stable across rounds and satisfies the verification threshold. This speculative path operates alongside the pipeline's ordinary confidence gate, which retains the high-confidence safe-commit rule.

A draft is never written into the committed sequence itself. Each active position is instead one of three states, listed in Table~\ref{tab:posstates}: committed, tentative, or masked, with tentative drafts held in a separate buffer. This separation matters to the pipeline, since the activation rule of Section~\ref{sec:pipeline} gauges a block's committed fraction by counting non-mask entries of the committed sequence, and routing drafts through it would inflate that count and open blocks too early.

Within each round, a \textit{re-drafting} step refreshes the drafts alongside commitment. Every still-masked position whose confidence reaches $\tau_d$ records its current prediction as a fresh draft, and any draft that failed verification is refreshed to the latest prediction. Both updates happen inside the same forward pass that performs the commit, which holds the amortized cost at one pass per round. The threshold $\tau_d$ is the procedure's main safety valve: it determines how reliable a prediction must be before it may act as context, and this protects later denoising from misleading drafts.

\subsection{Verification}
A draft provides provisional context before commitment, and each decoding round applies two commit rules before re-drafting.

\noindent\textbf{Safe-commit.} Every masked position whose confidence reaches $\tau$ is committed using the standard confidence-based criterion. Tentative context alone does not trigger commitment; a token must satisfy either this confidence criterion or the verification rule below before being finalized, apart from the progress-guaranteeing fallback.

\noindent\textbf{Verify-commit.} A tentative position is committed when its prediction is unchanged from the draft, $\hat{x}_i = \tilde{x}_i$, and its confidence reaches a verify threshold $\tau_v$. This rule admits tokens that stay consistent across rounds, including sub-$\tau$ tokens that the confidence gate alone would discard round after round, and it rejects tokens that are only briefly confident.

\noindent\textbf{Fallback.}
If neither commit path selects a token, \method{} commits the highest-confidence unresolved position, following the fallback used by the original decoder. This guarantees progress even when no prediction reaches either commit threshold. In our fixed-length setting, decoding terminates once all $L$ positions are committed and all blocks are frozen. 

The effect of this second path appears in commit lag, the number of consecutive steps a position predicts the same token before finalization. Fig.~\ref{fig:draft_analysis} shows that draft-and-verify reduces average lag from 11.7 to 8.9 (23.9\%) on held-out GSM8K and increases token confidence earlier in decoding. Both suggest that provisional context helps unresolved predictions stabilize and commit sooner.

\section{Experiments}
\label{sec:experiments}

\begin{table*}[!t]
\centering
\small
\setlength{\tabcolsep}{0pt}

\caption{
Main results on GSM8K, Minerva MATH, and MBPP. Cached variants use DualCache (Fast-dLLM), DSB's cache (DSB), and prefix KV cache (\method{}). Spd.\ denotes TPS speedup over uncached Original. For Acc., \textbf{bold} and \underline{underline} mark the top two among Original and cached variants (ties included). For NFE, TPS, and Spd., \textbf{bold} and \underline{underline} mark the best across all methods.
}
\label{tab:main-results}

\begin{tabular}{w{l}{71.5pt}*{12}{w{c}{31.95pt}}}
\toprule
\multirow{2}{*}{Method}
& \multicolumn{4}{c}{GSM8K}
& \multicolumn{4}{c}{Minerva MATH}
& \multicolumn{4}{c}{MBPP} \\
\cmidrule(lr){2-5} \cmidrule(lr){6-9} \cmidrule(l){10-13}
& Acc.\,(\%) & NFE\,$\downarrow$ & TPS\,$\uparrow$ & Spd.\,$\uparrow$
& Acc.\,(\%) & NFE\,$\downarrow$ & TPS\,$\uparrow$ & Spd.\,$\uparrow$
& Acc.\,(\%) & NFE\,$\downarrow$ & TPS\,$\uparrow$ & Spd.\,$\uparrow$ \\
\midrule

\multicolumn{13}{c}{\textbf{LLaDA-8B-Instruct}} \\
\midrule

Original
& 76.8 & 256.0 & 16.8 & 1.00$\times$
& \underline{33.2} & 256.0 & 24.1 & 1.00$\times$
& \underline{40.2} & 256.0 & 16.2 & 1.00$\times$ \\

\groupsep

Fast-dLLM
& 77.5 & 78.2 & 54.7 & 3.25$\times$
& 33.0 & 97.1 & 63.5 & 2.63$\times$
& 40.6 & 70.1 & 59.1 & 3.65$\times$ \\

DSB
& 77.7 & 74.5 & 57.4 & 3.41$\times$
& 33.1 & 93.6 & 65.6 & 2.72$\times$
& 41.8 & 67.2 & 60.2 & 3.72$\times$ \\

\rowcolor{methodbg}
\textbf{\method{}}
& 76.7 & \textbf{61.8} & 71.7 & 4.26$\times$
& 32.5 & \textbf{77.4} & 81.8 & 3.40$\times$
& 41.2 & \textbf{53.1} & 75.8 & 4.68$\times$ \\

\groupsep

Fast-dLLM + cache
& 77.3 & 85.3 & 90.7 & 5.39$\times$
& 32.7 & 107.3 & 80.0 & 3.32$\times$
& 37.8 & 70.2 & 76.0 & 4.70$\times$ \\

DSB + cache
& \textbf{78.4} & 78.4 & 100.8 & 5.99$\times$
& \textbf{33.5} & 98.0 & 88.5 & 3.67$\times$
& \textbf{42.6} & 65.8 & 79.7 & 4.93$\times$ \\

\rowcolor{methodbg}
\textbf{\method{} + cache}
& \underline{78.1} & 64.1 & \textbf{132.4} & \textbf{7.87$\times$}
& 32.9 & 80.0 & \textbf{122.9} & \textbf{5.10$\times$}
& 39.0 & 53.7 & \textbf{110.9} & \textbf{6.85$\times$} \\

\midrule

\multicolumn{13}{c}{\textbf{LLaDA-1.5}} \\
\midrule

Original
& \underline{80.3} & 256.0 & 15.7 & 1.00$\times$
& \underline{33.5} & 256.0 & 21.3 & 1.00$\times$
& \textbf{39.2} & 256.0 & 5.9 & 1.00$\times$ \\

\groupsep

Fast-dLLM
& 80.8 & 75.3 & 52.7 & 3.36$\times$
& 33.6 & 95.6 & 56.8 & 2.66$\times$
& 39.2 & 42.0 & 34.4 & 5.80$\times$ \\

DSB
& 81.0 & 71.7 & 55.6 & 3.54$\times$
& 34.2 & 94.4 & 57.0 & 2.67$\times$
& 34.8 & 42.0 & 32.9 & 5.55$\times$ \\

\rowcolor{methodbg}
\textbf{\method{}}
& 79.2 & \textbf{59.7} & 69.2 & 4.41$\times$
& 32.2 & \textbf{76.1} & 72.6 & 3.40$\times$
& 39.8 & 32.4 & 44.4 & 7.50$\times$ \\

\groupsep

Fast-dLLM + cache
& \textbf{80.6} & 84.2 & 84.1 & 5.36$\times$
& 32.4 & 105.5 & 70.9 & 3.32$\times$
& 34.2 & 43.9 & 48.6 & 8.21$\times$ \\

DSB + cache
& \textbf{80.6} & 76.2 & 92.9 & 5.92$\times$
& \textbf{34.0} & 98.1 & 76.4 & 3.58$\times$
& 37.0 & 38.6 & 47.9 & 8.08$\times$ \\

\rowcolor{methodbg}
\textbf{\method{} + cache}
& \textbf{80.6} & 62.1 & \textbf{128.6} & \textbf{8.20$\times$}
& 32.4 & 78.7 & \textbf{111.0} & \textbf{5.20$\times$}
& \underline{38.6} & \textbf{32.2} & \textbf{69.7} & \textbf{11.77$\times$} \\

\midrule

\multicolumn{13}{c}{\textbf{Dream-v0-Base-7B}} \\
\midrule

Original
& \underline{75.0} & 256.0 & 10.0 & 1.00$\times$
& \textbf{34.8} & 256.0 & 18.1 & 1.00$\times$
& \underline{53.0} & 256.0 & 5.3 & 1.00$\times$ \\

\groupsep

Fast-dLLM
& 75.4 & 152.6 & 16.7 & 1.67$\times$
& 34.4 & 102.7 & 44.6 & 2.46$\times$
& 53.8 & 90.3 & 14.6 & 2.74$\times$ \\

DSB
& 75.4 & 161.4 & 15.5 & 1.56$\times$
& 34.5 & 98.9 & 45.5 & 2.51$\times$
& 53.8 & 93.7 & 13.8 & 2.60$\times$ \\

\rowcolor{methodbg}
\textbf{\method{}}
& 73.6 & \textbf{121.1} & 21.3 & 2.14$\times$
& 32.9 & \textbf{89.5} & 52.4 & 2.89$\times$
& 53.2 & \textbf{80.0} & 16.7 & 3.14$\times$ \\

\groupsep

Fast-dLLM + cache
& \textbf{75.9} & 162.3 & 30.9 & 3.10$\times$
& \underline{33.3} & 121.8 & 57.2 & 3.15$\times$
& \underline{53.0} & 111.5 & 17.1 & 3.23$\times$ \\

DSB + cache
& 72.2 & 160.1 & 29.2 & 2.93$\times$
& 33.2 & 113.3 & 58.2 & 3.21$\times$
& \textbf{55.0} & 101.6 & 18.0 & 3.38$\times$ \\

\rowcolor{methodbg}
\textbf{\method{} + cache}
& 74.1 & 122.4 & \textbf{48.9} & \textbf{4.91$\times$}
& 32.0 & 95.5 & \textbf{90.3} & \textbf{4.98$\times$}
& 51.8 & 85.8 & \textbf{27.1} & \textbf{5.11$\times$} \\

\midrule

\multicolumn{13}{c}{\textbf{Dream-v0-Instruct-7B}} \\
\midrule

Original
& \textbf{79.7} & 256.0 & 12.3 & 1.00$\times$
& \textbf{37.8} & 256.0 & 20.6 & 1.00$\times$
& \underline{53.6} & 256.0 & 5.8 & 1.00$\times$ \\

\groupsep

Fast-dLLM
& 78.8 & 63.2 & 48.5 & 3.93$\times$
& 37.7 & 87.1 & 58.9 & 2.86$\times$
& 55.6 & \textbf{34.1} & 39.6 & 6.77$\times$ \\

DSB
& 79.3 & 81.1 & 37.0 & 3.00$\times$
& 37.7 & 90.3 & 56.2 & 2.73$\times$
& 50.6 & 45.2 & 28.0 & 4.80$\times$ \\

\rowcolor{methodbg}
\textbf{\method{}}
& 77.3 & \textbf{59.8} & 52.6 & 4.26$\times$
& 36.2 & \textbf{78.8} & 65.0 & 3.16$\times$
& 55.4 & 35.0 & 40.8 & 6.98$\times$ \\

\groupsep

Fast-dLLM + cache
& 76.6 & 71.8 & 73.3 & 5.94$\times$
& \underline{37.7} & 97.9 & 79.2 & 3.85$\times$
& 53.4 & 43.5 & 45.1 & 7.71$\times$ \\

DSB + cache
& \underline{77.8} & 67.5 & 75.4 & 6.11$\times$
& \textbf{37.8} & 90.2 & 81.5 & 3.96$\times$
& \textbf{57.4} & 36.0 & 50.5 & 8.63$\times$ \\

\rowcolor{methodbg}
\textbf{\method{} + cache}
& 77.6 & 60.5 & \textbf{97.8} & \textbf{7.93$\times$}
& 35.6 & 79.7 & \textbf{115.3} & \textbf{5.61$\times$}
& 52.6 & 35.2 & \textbf{56.9} & \textbf{9.73$\times$} \\

\bottomrule
\end{tabular}
\end{table*}

\paragraph{Models and benchmarks.}

We test \method{} on four diffusion language models from two widely used open dLLM families: LLaDA-8B-Instruct~\citep{nie2026large}, LLaDA-1.5~\citep{zhu2025llada15}, Dream-v0-Base-7B, and Dream-v0-Instruct-7B~\citep{ye2025dream}. We evaluate mathematical reasoning on GSM8K~\citep{cobbe2021training} and Minerva MATH~\citep{hendrycks2021math,lewkowycz2022solving}, and code generation on MBPP~\citep{austin2021program}. GSM8K uses 5-shot exact match, Minerva MATH uses 4-shot \texttt{math\_verify}, and MBPP uses 3-shot \texttt{pass@1}, all evaluated with \texttt{lm-eval}~\citep{biderman2024lessons}.

\paragraph{Baselines.}

We compare against three decoders. \textit{Original} is standard block-wise decoding with the full denoising budget and serves as the reference for speedup. \textit{Fast-dLLM}~\citep{wu2025fastdllm} commits every position whose confidence exceeds a threshold, and \textit{DSB}~\citep{luo2026dsb} is a training-free method that relaxes fixed block boundaries. Each method is evaluated without KV caching, to isolate reductions in number of forward evaluations, and with its own cache: DualCache for Fast-dLLM, the DSB cache for DSB, and Fast-dLLM's block-wise prefix KV cache for \method{}.

\paragraph{Metrics.}

We report accuracy, the average number of model forward evaluations (NFE) per problem, and throughput in tokens per second (TPS). TPS counts the generated answer tokens, excluding EOS padding, per second of generation time. Speedup (Spd.) is the unrounded TPS ratio relative to \textit{Original} without caching.


\paragraph{Implementation details.}

All experiments use a single NVIDIA H200 GPU, batch size 1, and generation length $256$ unless noted otherwise. We report mean TPS over three runs. With caching, \method{} adopts Fast-dLLM's block-wise approximate prefix KV cache~\citep{wu2025fastdllm}: prompt and frozen-block KV states are cached and refreshed when a block freezes, while active and pending blocks are recomputed each forward pass. Full configurations appear in Appendix~\ref{app:configs}.

\subsection{Quantitative Results}
\label{sec:main-results}

\begin{table}[t]
\centering
\small
\setlength{\tabcolsep}{4pt}
\caption{
Average accuracy change and geometric-mean speedup over original across 4 models and 3 benchmarks with KV caching. See Appendix Table~\ref{tab:appendix-average} for per-benchmark results.
}
\label{tab:main-average}

\begin{tabular}{lrr}
\toprule
Method & $\Delta$Acc.\ (pp) & Spd. $\uparrow$ \\
\midrule
Fast-dLLM + cache & $-$1.02 & 4.50$\times$ \\
DSB + cache & \textbf{+0.20} & 4.72$\times$ \\
\rowcolor{methodbg}
\textbf{\method{} + cache} & $-$0.98 & \textbf{6.65$\times$} \\
\bottomrule
\end{tabular}
\end{table}


Table~\ref{tab:main-results} shows that \method{} improves decoding efficiency across four models and three benchmarks. Without caching, it achieves 2.14$\times$--7.50$\times$ speedups over \textit{Original}, outperforming Fast-dLLM and DSB in TPS across all 12 settings (1.03$\times$--1.31$\times$ over Fast-dLLM), with lower NFE in 11 and 12 settings, respectively. With caching, \method{} achieves the highest TPS and lowest NFE among cached decoders in all 12 settings, outperforming Fast-dLLM with DualCache by 1.26--1.58$\times$ despite caching only the prefix. Table~\ref{tab:main-average} shows that despite task-dependent accuracy fluctuations across decoding policies, \method{} maintains average accuracy within one percentage point of \textit{Original} across all 12 cached settings, while outperforming Fast-dLLM in both average accuracy and speedup. Appendix~\ref{app:longer} confirms that efficiency gains persist at length 512.

\subsection{Ablation Studies}
\label{sec:ablations}

We ablate both components of \method{} on LLaDA-8B-Instruct with GSM8K, varying one hyperparameter at a time with the others at their defaults. Setting $K=1$ removes cross-block pipelining, and any $\tau_d \geq \tau$ removes tentative drafting, since every position reaching $\tau_d$ is already safe-committed.

\begin{table}[t]
\centering
\small
\setlength{\tabcolsep}{4pt}
\caption{Effect of pipeline depth $K$ (LLaDA-8B-Instruct, GSM8K). Larger $K$ raises throughput at the cost of accuracy. Spd.\ is the TPS speedup over Original.}
\label{tab:ablation-k}

\begin{tabular}{lrrrr}
\toprule
$K$ & Acc. (\%) & NFE $\downarrow$ & TPS $\uparrow$ & Spd. $\uparrow$ \\
\midrule
1 (no pipelining) & 78.2 & 72.4 & 61.0 & 3.63$\times$ \\
2 & 76.7 & 61.8 & 71.7 & 4.26$\times$ \\
3 & 75.6 & 46.8 & 95.1 & 5.66$\times$ \\
4 & 74.3 & 41.8 & 105.9 & 6.30$\times$ \\
\bottomrule
\end{tabular}
\end{table}



\paragraph{Effect of pipeline depth.}

Table~\ref{tab:ablation-k} shows the effect of varying the pipeline depth $K$: raising $K$ from 1 to 2 increases the speedup from 3.63$\times$ to 4.26$\times$ at a cost of 1.5 points. Larger $K$ raises throughput further, up to 6.30$\times$ at $K=4$, but with a steeper accuracy cost, so we use $K=2$.

\begin{table}[t]
\centering
\small
\setlength{\tabcolsep}{4pt}
\caption{Effect of draft threshold $\tau_d$ (LLaDA-8B-Instruct, GSM8K). Lower $\tau_d$ raises throughput, but overly low values make accuracy unstable.}
\label{tab:ablation-draft-threshold}

\begin{tabular}{lrrrr}
\toprule
$\tau_d$ & Acc. (\%) & NFE $\downarrow$ & TPS $\uparrow$ & Spd. $\uparrow$ \\
\midrule
0.6 & 75.2 & 42.1 & 104.3 & 6.20$\times$ \\
0.7 & 77.1 & 48.7 & 90.9 & 5.41$\times$ \\
0.8 & 76.5 & 57.4 & 77.2 & 4.59$\times$ \\
0.85 & 76.7 & 61.8 & 71.7 & 4.26$\times$ \\
$\geq \tau$ (no drafting) & 77.4 & 66.8 & 66.2 & 3.94$\times$ \\

\bottomrule
\end{tabular}
\end{table}



\paragraph{Effect of draft threshold.}

Table~\ref{tab:ablation-draft-threshold} shows the effect of varying the draft threshold $\tau_d$: lowering $\tau_d$ from $\tau$ (no drafting) to 0.85 raises the speedup from 3.94$\times$ to 4.26$\times$. Lower thresholds admit less reliable drafts, making accuracy unstable before it drops to 75.2\% at $\tau_d=0.6$, so we keep $\tau_d=0.85$ as the default for both model families.

\subsection{Validating the Verification Mechanism}
\label{sec:verification-diagnostic}

To test whether verify-commit simply copies tentative context rather than verifying it, we replace each GSM8K draft with a random vocabulary token in a measurement-only forward pass that does not affect the actual decoding process.

\begin{table}[t]
\centering
\small
\caption{Verification diagnostic on GSM8K. Random token replacements cause most drafts to be rejected.}
\label{tab:verification-diagnostic}
\begin{tabular}{lrrr}
\toprule
Slot content & Positions & Admitted & Rejected \\
\midrule
Genuine draft & 37,091 & 97.31\% & 2.69\% \\
Random token  & 37,091 & 16.46\% & \textbf{83.54\%} \\
\bottomrule
\end{tabular}
\end{table}

Table~\ref{tab:verification-diagnostic} shows that 83.54\% of random drafts are rejected, versus 2.69\% of genuine drafts, indicating that verification favors drafts consistent with the model's prediction rather than merely copying the tentative input context.




\section{Conclusion}


We introduce Tentative Context Pipelining (\method{}), a training-free decoder that decouples contextual utility from commitment readiness through cross-block pipelining and draft-and-verify. By reusing stable but uncommitted predictions as provisional context, \method{} reduces number of forward evaluations by up to 87\% and achieves up to 7.50$\times$ and 11.77$\times$ speedups over uncached standard block-wise decoding without and with KV caching, respectively. Across four dLLMs and three benchmarks, \method{} outperforms Fast-dLLM and DSB in throughput across all 12 settings with and without caching, achieving a 6.65$\times$ geometric-mean cached speedup while keeping average accuracy within one percentage point of the standard decoder.

\section*{Limitations}

\method{} remains unevaluated on open-ended and long-form generation benchmarks, where higher output entropy may reduce draft stability and longer sequences may increase error propagation. Future work will study broader tasks, longer generations, and cross-domain draft stability.


\bibliography{ref}

\clearpage
\appendix



\section{Experimental Details and Additional Results}
\label{app:configs}

This appendix provides additional experimental details and analyses. Unless otherwise specified, experiments use a generation length of $256$ and batch size 1. For \method{}, one configuration is shared within each model family and fixed across GSM8K, Minerva MATH, and MBPP. All benchmarks are in English and cover mathematical reasoning or Python code generation.

\paragraph{Implementation and licenses.}
Our implementation builds on existing open-source implementations of LLaDA, Dream, and Fast-dLLM under their respective licenses, with all required notices retained. Model weights and benchmark datasets are not redistributed.

\subsection{Hyperparameter Notation}

Table~\ref{tab:param-notation} summarizes the decoding parameters used throughout the experiments.

\begin{table}[htbp]
\centering
\small
\setlength{\tabcolsep}{4pt}
\renewcommand{\arraystretch}{1.05}
\caption{Notation of decoding parameters used in \method{}.}
\label{tab:param-notation}
\begin{tabularx}{\columnwidth}{
    @{}
    >{\centering\arraybackslash}p{0.18\columnwidth}
    >{\raggedright\arraybackslash}X
    @{}
}
\toprule
Parameter & Meaning \\
\midrule
$b$        & Block size. \\
$K$        & Max number of concurrent active blocks. \\
$\alpha$   & Threshold for opening the next block. \\
$\tau$     & Commit threshold for safe-commit. \\
$\tau_d$   & Threshold for admitting tentative drafts. \\
$\tau_v$   & Threshold for committing stable tentative drafts. \\
\bottomrule
\end{tabularx}
\end{table}

\subsection{Hyperparameter Settings}

Table~\ref{tab:model-configs} reports the decoding configurations used for \method{}. We use one shared configuration for the two LLaDA models and another shared configuration for the two Dream models, reflecting differences in their confidence distributions and decoding behavior. Importantly, these hyperparameters are not tuned separately for each benchmark: once selected, the same model-family-level configuration is reused across GSM8K, Minerva MATH, and MBPP.

\begin{table}[htbp]
\centering
\small
\caption{Model-family-specific decoding configuration for \method{}.}
\label{tab:model-configs}
\resizebox{\columnwidth}{!}{%
\begin{tabular}{lcccccc}
\toprule
Model & $b$ & $K$ & $\alpha$ & $\tau$ & $\tau_d$ & $\tau_v$ \\
\midrule
LLaDA-8B-Instruct      & 32 & 2 & 0.80 & 0.90 & 0.85 & 0.80 \\
LLaDA-1.5              & 32 & 2 & 0.80 & 0.90 & 0.85 & 0.80 \\
Dream-v0-Base-7B       & 16 & 2 & 0.85 & 0.90 & 0.85 & 0.70 \\
Dream-v0-Instruct-7B   & 16 & 2 & 0.85 & 0.90 & 0.85 & 0.70 \\
\bottomrule
\end{tabular}%
}
\vspace{-10pt}
\end{table}

\paragraph{Baseline and caching configurations.}
All methods use greedy decoding. Original uses block size 32 and commits one position per forward pass, with low-confidence remasking for LLaDA and entropy-based ordering for Dream. Fast-dLLM and DSB use a confidence threshold of 0.9 and an initial block size of 32. In the cached setting, Fast-dLLM uses DualCache, DSB uses its official cache with a prefix window of 24 for LLaDA and 4 for Dream, and \method{} uses the block-wise prefix KV cache of Fast-dLLM, refreshed whenever a block is frozen.

\subsection{Average Metric Per Dataset}
Table~\ref{tab:appendix-average} shows the per-benchmark average accuracy change and
geometric-mean speedup relative to \textit{Original}, over 4 models with caching enabled.
\begin{table}[htbp]
\centering
\small
\setlength{\tabcolsep}{4pt}
\caption{
Average accuracy change and geometric-mean speedup relative to \textit{Original}.
}
\label{tab:appendix-average}

\begin{tabular}{lrr}
\toprule
Method & $\Delta$Acc.\ (\%) & Spd. $\uparrow$ \\
\midrule

\multicolumn{3}{c}{\textbf{GSM8K}} \\
\midrule
Fast-dLLM + cache & \textbf{$-$0.35} & 4.80$\times$ \\
DSB + cache & $-$0.70 & 5.02$\times$ \\
\rowcolor{methodbg}
\textbf{\method{} + cache} & \textbf{$-$0.35} & \textbf{7.08$\times$} \\
\midrule

\multicolumn{3}{c}{\textbf{Minerva MATH}} \\
\midrule
Fast-dLLM + cache & $-$0.80 & 3.40$\times$ \\
DSB + cache & \textbf{$-$0.20} & 3.59$\times$ \\
\rowcolor{methodbg}
\textbf{\method{} + cache} & $-$1.60 & \textbf{5.22$\times$} \\
\midrule

\multicolumn{3}{c}{\textbf{MBPP}} \\
\midrule
Fast-dLLM + cache & $-$1.90 & 5.57$\times$ \\
DSB + cache & \textbf{+1.50} & 5.84$\times$ \\
\rowcolor{methodbg}
\textbf{\method{} + cache} & $-$1.00 & \textbf{7.96$\times$} \\

\bottomrule
\end{tabular}
\end{table}

\subsection{Longer Generation Length}
\label{app:longer}

We further evaluate decoding with a generation length of 512 on GSM8K, using the same \method{} configurations as in the main experiments without retuning. KV caching remains disabled and batch size is 1. As shown in Table~\ref{tab:len512}, \method{} retains substantial efficiency gains at the longer generation length, achieving speedups of 7.18$\times$ and 8.44$\times$ over Original on LLaDA-8B-Instruct and LLaDA-1.5, respectively. Both exceed their corresponding speedups at length 256, and \method{} achieves the highest throughput among all compared methods on both models, demonstrating that its efficiency gains extend to longer generation lengths.

\begin{table}[htbp]
\centering
\small
\setlength{\tabcolsep}{4pt}
\caption{Results on GSM8K with generation length $512$, without KV caching. \method{} uses the length-$256$ configurations without retuning. Spd.\ is the TPS speedup over Original, computed from unrounded TPS.}
\label{tab:len512}
\begin{tabular}{lrrrr}
\toprule
Method & Acc. (\%) & NFE $\downarrow$ & TPS $\uparrow$ & Spd. $\uparrow$ \\
\midrule
\multicolumn{5}{c}{\textbf{LLaDA-8B-Instruct}} \\
\midrule
Original   & 77.6 & 512.0 & 8.0  & 1.00$\times$ \\
Fast-dLLM  & 76.9 & 97.0  & 44.1 & 5.50$\times$ \\
DSB        & 77.7 & 97.8  & 43.2 & 5.39$\times$ \\
\rowcolor{methodbg}
\textbf{\method{}} & 76.2 & \textbf{75.6} & \textbf{57.5} & \textbf{7.18$\times$} \\
\midrule
\multicolumn{5}{c}{\textbf{LLaDA-1.5}} \\
\midrule
Original   & 81.6 & 512.0 & 6.1  & 1.00$\times$ \\
Fast-dLLM  & 83.0 & 80.3  & 40.0 & 6.57$\times$ \\
DSB        & 82.8 & 80.8  & 39.1 & 6.43$\times$ \\
\rowcolor{methodbg}
\textbf{\method{}} & 82.5 & \textbf{64.3} & \textbf{51.4} & \textbf{8.44$\times$} \\
\bottomrule
\end{tabular}
\end{table}
\subsection{Peak GPU Memory}
\label{app:peak-memory}

We additionally measure peak allocated GPU memory on 100 GSM8K samples with batch size 1, generation length 256, and KV caching disabled.

\begin{table}[htbp]
\centering
\small
\setlength{\tabcolsep}{4pt}
\caption{Peak allocated GPU memory (GiB) on GSM8K without KV caching. Original and Fast-dLLM have identical peak memory in this setting.}
\label{tab:peak-memory}
\begin{tabular}{lrr}
\toprule
Model & Original / Fast-dLLM & \method{} \\
\midrule
LLaDA-8B-Instruct    & 18.53 & \textbf{16.17} \\
LLaDA-1.5            & 18.51 & \textbf{16.16} \\
Dream-v0-Base-7B     & \textbf{15.21} & 16.13 \\
Dream-v0-Instruct-7B & \textbf{15.22} & 16.15 \\
\bottomrule
\end{tabular}
\end{table}

As shown in Table~\ref{tab:peak-memory}, \method{} introduces no substantial additional peak-memory overhead. All configurations remain below 19\,GiB: \method{} uses less peak memory on LLaDA and incurs less than 1\,GiB additional memory on Dream.

\end{document}